\documentclass{article}

\usepackage{PRIMEarxiv}

\usepackage[utf8]{inputenc} % allow utf-8 input
\usepackage[T1]{fontenc}    % use 8-bit T1 fonts
\usepackage{hyperref}       % hyperlinks
\usepackage{url}            % simple URL typesetting
\usepackage{booktabs}       % professional-quality tables
\usepackage{amsfonts}       % blackboard math symbols
\usepackage{nicefrac}       % compact symbols for 1/2, etc.
\usepackage{microtype}      % microtypography
\usepackage{lipsum}
\usepackage{fancyhdr}       % header
\usepackage{graphicx}       % graphics
\usepackage[table,xcdraw]{xcolor}
\usepackage{caption}
\usepackage{amsmath}
\graphicspath{{./}}

\usepackage{esint}

\title{Toward Developing Machine-Learning-Aided Tools for the Thermomechanical Monitoring of Nuclear Reactor Components
}

\author{
  Luiz Aldeia Machado, Victor Coppo Leite, Elia Merzari, Arthur Motta  \\
  Ken and Mary Alice Lindquist Department of Nuclear Engineering\\
  Pennsylvania State University,\\
  University Park, Pennsylvania, 16802\\
  \texttt{lca5209@psu.edu} \\
  \And
  Roberto Ponciroli, Lander Ibarra\\
  Nuclear Science and Engineering Division \\
  Argonne National Laboratory\\
  Lemont, Illinois, 60439
  \AND
  Lise Charlot\\
  Idaho National Laboratory\\
  Idaho Falls, ID, 83415
}

\begin{document}
\maketitle

\begin{abstract}
Proactive maintenance strategies, such as Predictive Maintenance (PdM), play an important role in the operation of Nuclear Power Plants (NPPs), particularly due to their capacity to reduce offline time by preventing unexpected shutdowns caused by component failures. 

In this work, we explore using a Convolutional Neural Network (CNN) architecture combined with a computational thermomechanical model to calculate the temperature, stress, and strain of a Pressurized Water Reactor (PWR) fuel rod during operation. This estimation relies on a limited number of temperature measurements from the cladding’s outer surface. This methodology can potentially aid in developing PdM tools for nuclear reactors by allowing the real-time monitoring of such systems.

The training, validation, and testing datasets were generated through coupled simulations involving BISON, a finite element-based nuclear fuel performance code, and the Multiphysics Object-Oriented Simulation Environment Thermal-Hydraulic Module (MOOSE-THM). We conducted eleven simulations, varying the peak linear heat generation rates. Of these, eight were used for training, two for validation, and one for testing.

The CNN was trained for over 1,000 epochs without signs of overfitting, achieving highly accurate temperature distribution predictions. Our thermomechanical model used these predictions to determine the stress and strain distribution within the fuel rod.
\end{abstract}

% keywords can be removed

\section{Introduction}
Implementing proactive maintenance strategies, such as Predictive Maintenance (PdM), can increase the reliability of Nuclear Power Plants (NPPs). PdM employs continuous monitoring and periodic inspections to identify potential issues before they lead to equipment failure, reducing Operation and Maintenance (O\&M) costs and enhancing overall reactor safety \cite{6246396, plants2014using}.

PdM can be implemented using either a probabilistic approach, which relies on statistical and reliability analysis, and the data is retrieved from the Computerized Maintenance Management Information System (CMMIS) software, or a deterministic approach, which leverages sensor data, data mining techniques, and artificial intelligence algorithms to predict component failures. The deterministic approach often leads to better maintenance cost optimization compared to the probabilistic method \cite{saley2022state}. However, the success of this approach relies on accurate data collection, which can be challenging in advanced reactor designs. Harsh operating conditions and intense radiation exposure may make sensor readings impractical in certain areas of the reactor vessel.

Ponciroli et al. \cite{cnn1} proposed a Convolutional Neural Network (CNN) architecture capable of estimating the spatial distribution of a field in regions where invasive monitoring approaches cannot be employed. Their algorithm formulates the monitoring problem of a physical quantity governed by the Helmholtz equation as a Boundary Value Problem (BVP), which is solved using the Boundary Element Method (BEM), enabling accurate spatial field reconstruction.

Building on this work, Leite et al. (2023-a) \cite{vic_htgr} adapted the architecture presented in \cite{cnn1} to determine the temperature distribution within the solid domain of a High-Temperature Gas Reactor (HTGR). This approach effectively detected hotspots resulting from a hypothetical channel blockage using a limited set of temperature readings taken from locations far from the hotspot. The CNN-based architecture was later utilized to reconstruct the coolant temperature distribution for a Molten Salt Fast Reactor (MSFR), as presented in \cite{vic_msfr}.

Aldeia Machado et al. \cite{aldeia2024temperature} expanded this methodology to reconstruct the temperature field on the vessel wall of an HTGR exposed to radiative heat transfer. They introduced a modified version of the original architecture that leverages, under certain conditions, the Newton Law of Cooling (NLC) to estimate the temperature normal derivative along the domain boundaries exposed to natural convection. This approach allows temperature field reconstruction even when the temperature gradient cannot be measured at the collocation points.  

This method enabled the direct reconstruction of the temperature field from thermocouple readings, allowing for the validation of the NLC-based CNN model. Additionally, using the reconstructed temperature field as input for a computational thermomechanical model,\cite{aldeia2024temperature} demonstrated the feasibility of estimating thermally induced strain on the vessel wall from thermocouple readings. These findings highlight the potential use of the approach for developing PdM strategies in nuclear reactors.

In this study, we utilize the NLC-based CNN model to reconstruct the temperature distribution of a Pressurized Water Reactor (PWR) fuel rod using four temperature measurements collected from the cladding outer surface. The reconstructed temperature field is then employed to estimate the stress and strain on the fuel rod, thereby broadening the applications of the work presented in \cite{aldeia2024temperature} to a new domain.

The data used for training, validation, and testing of the CNN model were obtained from a coupled simulation utilizing BISON, a finite element-based nuclear fuel performance code, and the Multiphysics Object-Oriented Simulation Environment Thermal-Hydraulic Module (MOOSE-THM). Eleven simulations were performed, each varying the peak linear heat generation rate. Specifically, eight simulations were used in the training dataset, two in the validation dataset and one in the testing dataset. The CNN was trained for over one thousand epochs without any signs of overfitting, resulting in a highly accurate temperature distribution prediction with an R-squared value of $0.999326$ compared to the computational model. An in-depth discussion about the coupled model and CNN architecture is provided in the Methodology section.

\section{METHODOLOGY}
\label{sec:methodology}

This section outlines the methodologies employed in this study. We introduce the computational model used to create the datasets for the CNN learning process. Subsequently, we describe the CNN model, the distribution of data among the training, validating, and testing datasets, and the assumptions our model makes regarding using the NLC-based CNN model.

\subsection{BISON Model}

In our computational model, BISON solves the temperature distribution within the fuel rodlet. BISON is also responsible for controlling the changes in the fuel material properties due to temperature, porosity, and composition changes during the reactor operation. Our model comprises a full-length PWR rod with $UO_2$ and Zircaloy-4 used as the fuel and cladding materials, respectively. We modeled the fuel rod as an RZ-axisymmetric problem thanks to its symmetry. The values of some geometric parameters are presented in Table \ref{tab:geo}. 

\begin{table}[!h]
\centering
\caption{Fuel rod geometric parameters.}
\label{tab:geo}
\begin{tabular}{ccccc}
\hline
$\mathbf{L_{fr}\,\,\,[m]}$ & $\mathbf{L_{f}\,\,\,[m]}$ & $\mathbf{R_{fo}\,\,\,[m]}$ & $\mathbf{R_{ci}\,\,\,[m]}$ & $\mathbf{R_{co}\,\,\,[m]}$ \\ \hline
3.876                    & 3.658                   & 0.004096                 & 0.0041786                & 0.0047506                \\ \hline
\end{tabular}
\end{table}

\noindent where $L_{fr}$ is the fuel rod length, $L_{f}$ is the fuel length, $R_{fo}$ is the fuel pellet outer radius, $R_{ci}$ is the cladding inner radius, and $R_{co}$ is the cladding outer radius. Table \ref{tab:bcs} summarizes the boundary conditions used in the BISON model.

\begin{table}[!h]
\caption{BISON model boundary conditions.}
\label{tab:bcs}
\begin{tabular}{lccc}
\hline
\textbf{Block}      & \textbf{Boundary} & \textbf{Type} & \textbf{Expression}                                                          \\ \hline
Pellet and Cladding & Centerline        & Neumann       & $q^{''}\left(r=0,z\right)=0$                                                 \\
Pellet and Cladding & Centerline        & Dirichlet     & $\delta_r\left(r=0,z\right)=0$                                               \\
Pellet              & Bottom            & Dirichlet     & $\delta_z\left(r,z=z_{pb}\right)=0$                                          \\
Cladding            & Bottom            & Dirichlet     & $\delta_z\left(r,z=0\right)=0$                                               \\
Cladding            & Bottom            & Neumann       & $q^{''}\left(r,z=0\right)=0$                                                 \\
Cladding            & Top               & Neumann       & $q^{''}\left(r,z=L_{fr}\right)$                                              \\
Cladding            & Outer Surface     & Robin         & $q^{''}\left(r=R_{co},z\right)=h\left(T\left(r=R_{co},z\right)-T_{\infty}\right)$ \\ \hline
\end{tabular}
\end{table}

where $\delta_r$ is the radial displacement, $\delta_z$ is the axial displacement, $z_{pb}$ is the axial fuel bottom coordinate, $h$ is the convective Heat Transfers Coefficient (HTC), and $T_{\infty}$ is the coolant temperature. 

\begin{figure}[!ht]
    \centering
    \captionsetup{justification=centering}
    \includegraphics[width=0.5\textwidth]{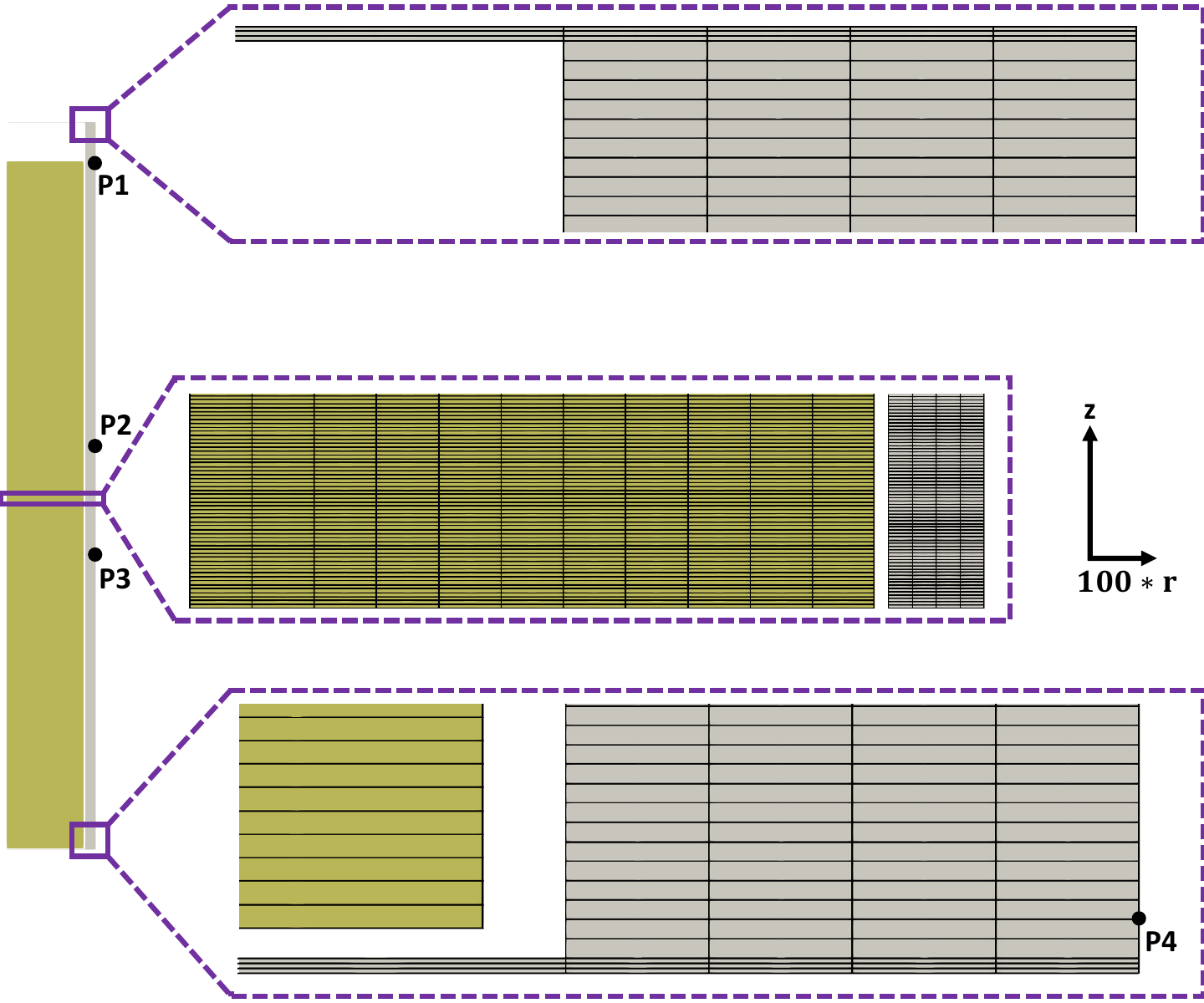}
    \caption{BISON model geometry $(100\times r)$.}
    \label{fig:bison_geo}
\end{figure}

Our fuel mesh consists of 11 nodes in the radial direction and 1,560 in the axial direction. For the cladding, these values are 4 nodes in the radial direction and 2,000 in the axial direction. Figure \ref{fig:bison_geo} illustrates the model geometry and indicates the approximate locations of the four points where we collected temperature measurements to perform the field reconstruction. Note that due to the difference between the axial and radial dimensions of our problem, the radial coordinates have been scaled up by two orders of magnitude to enhance the visualization of the geometry.

\noindent The fuel's initial enrichment was 4.5at\%, with thermal conductivity and fission gas release modeled using the NFIR (Nuclear Fuel Industry Research) correlation \cite{nfir} and SIFGRS (Simple Integrated Fission Gas Release and Swelling) model \cite{bison_manual}, respectively. We considered a sinusoidal heat generation rate, given by (Eq. \ref{eq:linear_heat}):

\begin{equation}
    q'(z) = q'_0\textbf{sin}\left(\pi\frac{z}{L_e}\right)
    \label{eq:linear_heat}
\end{equation}

\noindent where $q'_0$ is the peak linear heat generation rate, and $L_e$ is the fuel rod length plus the extrapolation length at the core bottom and top boundaries. The energy balance in BISON is given by (Eq. \ref{eq:bison_energy}) \cite{bison_manual}:

\begin{equation}
    \rho C_{p} \frac{\partial T}{\partial t} + \mathbf{\nabla}\cdot \mathbf{q^{''}} - e_f \dot{F} = 0,
    \label{eq:bison_energy}
\end{equation}

where T is the temperature, $\rho$ and $C_p$ are, respectively, the density and specific heat. $\dot{F}$ is the volumetric fission rate and $e_f$ is the energy released per fission event, and $q^{''}$ is the heat flux.

\subsection{MOOSE-THM Model}

 The MOOSE-THM geometry consists of a one-dimensional channel divided into 2,000 nodes. We prescribed the channel inlet temperature, $T_{in}$,  as 583.15 [K], the outlet pressure, $P_{out}$, as 15.51 [MPa], and the mass flux, $G$, as 3244.04 \text{$[Kg/s.m^2]$}. MOOSE-THM uses the following set of area-averaged balance equations  \cite{THM}:

For the mass balance:

\begin{equation}
        \frac{\partial}{\partial t}A\langle\rho\rangle_{A} + \frac{\partial}{\partial x}A\langle\rho U\rangle_{A} = 0
    \label{eq:thm_mass_balance}
\end{equation}

For the momentum balance:

\begin{equation}
    \frac{\partial}{\partial t}A\langle\rho U\rangle_A + \frac{\partial}{\partial x}A(\langle\rho U^2\rangle_A + \langle p\rangle_A) = -\Tilde{p}\frac{\partial A}{\partial x} - F_{friction}
    \label{eq:thm_momentum}
\end{equation}

For the total energy balance:

\begin{equation}
   \frac{\partial}{\partial t}A\langle\rho E\rangle_A + \frac{\partial}{\partial x}A(\langle\rho E\rangle_A + \langle pU\rangle_A) = -\Tilde{p}\frac{\partial A}{\partial t} +Q_{wall}
    \label{eq:thm_energy}
\end{equation}

where $U = \mathbf{U}\cdot\mathbf{\hat{n}_x}$ is the component of velocity in the flow direction, $\Tilde{p}$ is the average pressure around the perimeter of the duct cross-section, $F_{friction}$ is the average wall shear stress, $E = e + \frac{\mathbf{U}\mathbf{U}}{2}$ is the specific total energy, $e$ is the specific internal energy, and $Q_{wall}$ is the average heat transfer rate between from the duct wall to the fluid.

The convective HTC is calculated inside the MOOSE-THM model using the Dittus-Boelter correlation. For the friction factor, we used the Cheng and Todreas correlation for an interior subchannel and turbulent flow \cite{todreas}.

The information exchange between the BISON and MOOSE-THM models is managed by the MOOSE MultiAPP System, which facilitates the solution of multiphysics and multiscale problems by employing coupled models \cite{multiphysics}. Through this system, BISON sends the temperature distribution of the cladding's outer wall to MOOSE-THM. MOOSE-THM then uses this information to calculate the coolant's temperature distribution, $T_{cool}(z)$, and the convective HTC, $h(z)$. These values are sent back to BISON, where they are incorporated into the Robin boundary condition applied to the cladding outer surface.

\subsection{Thermomechanical Model}
\label{sec:thermomechanical}

Some failure criteria developed to determine the Pellet-Cladding Mechanical Interaction (PCMI) failure during accident conditions, such as the Nantes Criterion, \cite{nantes}, rely on the knowledge of the hoop strain to determine if the system failed or not due to PCMI. This work utilized the MOOSE Solid Mechanics Module (MOOSE-SMM) to develop two thermomechanical models for calculating the maximum cladding hoop strain in the studied example. 

The first model, referred to as the \textit{reference model}, refers to a PWR fuel rod with Zircaloy-4 cladding and $UO_2$ fuel. This rod is irradiated to a burnup of 16.7 MWd/KgU, reaching an average coolant, cladding, and fuel temperature of 598 $\text{[K]}$, 615 $\text{[K]}$, and 847 $\text{[K]}$, respectively. Under these conditions, the reference model in BISON calculates a cladding oxide thickness of 25.4 $[\mu m]$ at that burnup. At the end of the irradiation process for the reference model, the gap between the pellet and the cladding was partially closed, with the PCMI happening around the fuel rod’s axial centerline.  

The reference model includes the following hoop strain components:

\begin{itemize}
    \item \textbf{Fuel thermal expansion: }accounting for the isotropic thermal expansion of the fuel.
    \item \textbf{Fuel Volumetric Swelling: } accounting for the strain resulting from the changes in the fuel pellet volume due to densification and fission products.
    \item \textbf{Cladding irradiation growth: }accounting for the cladding irradiation-induced growth.
    \item \textbf{Cladding thermal expansion: }accounting for the anisotropic thermal expansion of the cladding tube.
    \item \textbf{Zircaloy creep strain: } accounting for the secondary hoop radiation creep and Limback-Anderson secondary and primary thermal creep.
\end{itemize}

The second thermomechanical model used in this work, which is referred to as the \textit{simplified model}, also models a PWR fuel rod with Zircaloy-4 cladding and $UO_2$ fuel but, differently from the reference model, here the radiation-induced effects within the fuel rod, such as the fuel swelling, fission gas release, and irradiation-induced growth, are not considered. This model takes the predicted temperature field from the CNN architecture as input and uses it to calculate the thermal-induced strain and stress, accounting for the fuel and cladding thermal expansion and the cladding thermal creep.

\subsection{Convolutional Neural Network Model}

In this study, we employed the CNN architecture developed by Ponciroli et al. \cite{cnn1} to reconstruct the temperature distribution of a PWR fuel rod based on a limited number of temperature measurements. Their approach involves solving the Helmholtz equation, a partial differential equation that describes various physical phenomena, notably heat diffusion in purely diffusive media. The Helmholtz equation is represented as follows (Eq. \ref{eq:helm}):

\begin{equation}
    \frac{\partial^2 u}{\partial \ n^2} = -\lambda^2u
    \label{eq:helm}
\end{equation}

\noindent where $\lambda^2$ is the wave number, a constant related to the frequency of the wave or the properties of the medium, and $u$ is the function of interest, which for our application represents the temperature.

The Helmholtz equation is commonly solved as a Boundary Value Problem (BVP) through the Boundary Element Method (BEM). This numerical approach relies on discretizing the Kirchhoff-Helmholtz (K-H) integral equation without source terms, given by (Eq.\ref{eq:KH}) \cite{KH_integral}:

\begin{equation}
    u(\mathbf{r}) = \oiint\limits_{\Sigma}\left(u(\mathbf{r'})\frac{\partial G(\mathbf{r},\mathbf{r'})}{\partial \mathbf{n}} - G(\mathbf{r},\mathbf{r'})\frac{\partial u(\mathbf{r'})}{\partial \mathbf{n}}\right)d\Sigma,
    \label{eq:KH}
\end{equation}

\noindent where $\Omega$ represents the monitored domain ($\mathbf{r} \in \Omega$), $\Sigma=\partial\Omega$ is the closed surface enclosing the domain  ($\mathbf{r'} \in \Sigma$), $\mathbf{n}$ is the outward unit normal vector at the domain boundary, and $G(\mathbf{r},\mathbf{r'})$ is the Green's function.

The CNN model described in \cite{cnn1} employs a supervised learning approach to evaluate the numerical expression of the unknown Green's function and its normal derivative. This network requires the field value $u(\mathbf{r'})$ and its normal derivative $\partial u(\mathbf{r'})/\partial \mathbf{n}$  to be known at the domain boundaries. However, \cite{aldeia2024temperature} introduced a way to relax some of these requirements by estimating the temperature normal derivative using Newton's Law of Cooling (NLC). They demonstrated the validity of the  NLC-based CNN architecture under the following assumptions:

\begin{enumerate}
    \item The reconstructed temperature field is governed only by the heat conduction equation.
    \item The convective heat transfer coefficient is small and nearly constant.
\end{enumerate}

These two characteristics make the guess convective HTC, $\eta$, in the NLC-based CNN architecture to be assimilated by the CNN as a scale factor, allowing them to perform the HGTR temperature field reconstruction using only thermocouple readings. 

In this work, the temperature distribution over the fuel rod is governed by the heat conduction equation, which agrees with requirement 1 of the NLC-based CNN architecture. However, our convective HTC is large and varies with the coolant temperature, which does not fulfill the second requirement. By applying the NLC-based CNN architecture to the fuel rod problem, we aim to demonstrate that such an approach can also be used in cases where the HTC is large and not spatially constant, expanding its applicability.

\begin{figure}[!ht]
    \centering \captionsetup{justification=centering}
    \includegraphics[width=\textwidth]{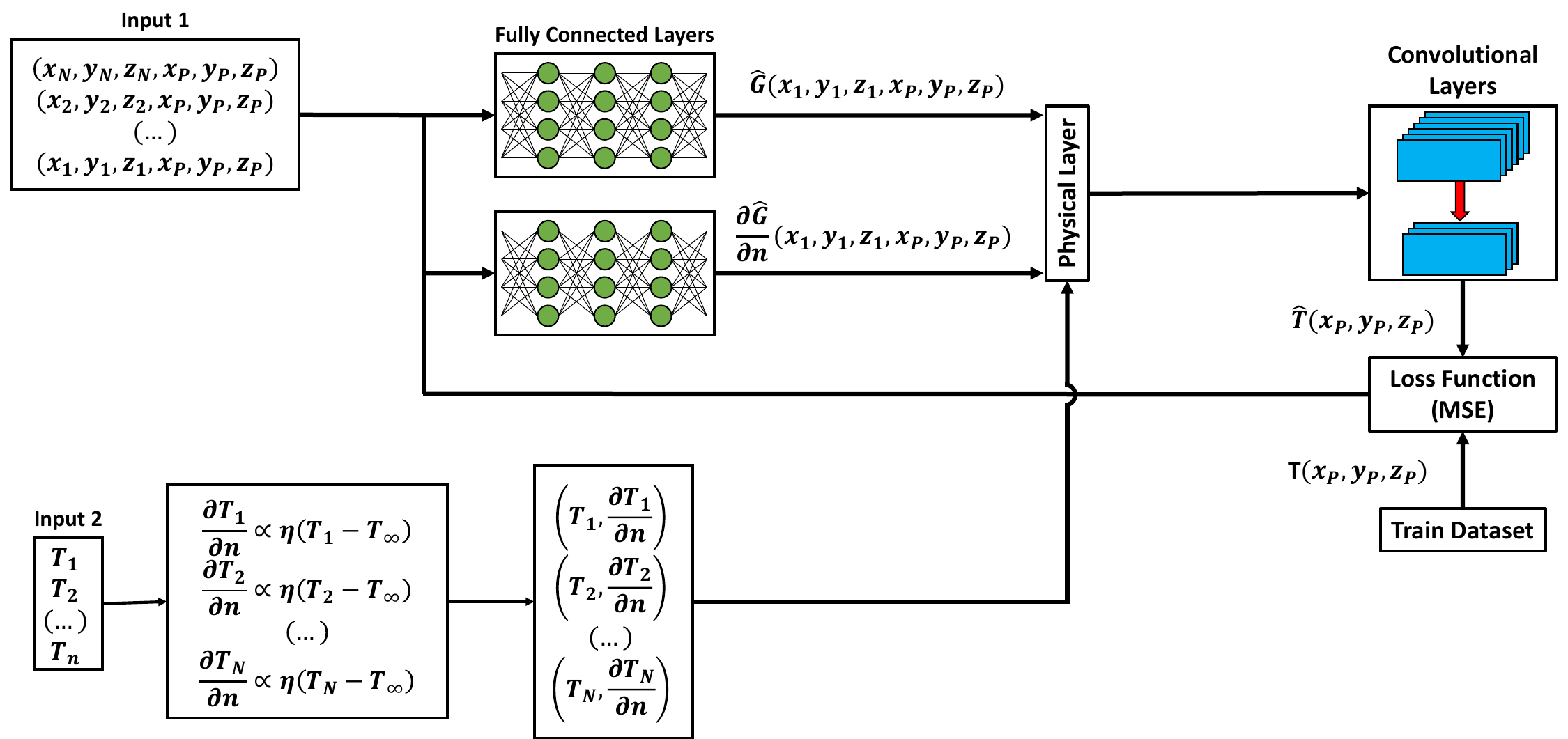}
    \caption{Graphical representation of the CNN architecture using the NLC approach, adapted from  \cite{aldeia2024temperature}.}
    \label{fig:cnn}
\end{figure}

Figure \ref{fig:cnn} shows the graphical representation of the network, which is composed of:

\begin{itemize}
    \item Two sets of fully connected layers: these layers are responsible for solving the Green's function and its normal derivative. 
    \item Physical layer: this layer will receive the outputs of the fully connected layers in addition to the information from input 2. It will compute the integrand of the K-H integral equation.
    \item Convolutional layer: this layer will integrate the outputs from the physical layer along the domain boundaries, estimating the temperature distribution at the interior points.
\end{itemize}

The learning process will minimize the loss function, which was defined as the Mean Square Error (MSE) (Eq.\ref{eq:mse})  between the estimated temperature values ($\hat{T}(x_i, y_i, z_i)$) and the measured value ($T(x_i, y_i, z_i)$) at the interior point coordinates \cite{cnn1}.

\begin{equation}
    MSE = \frac{1}{N_P}\sum_{i=1}^{N_P}\left[\hat{T}(x_i, y_i, z_i) - T(x_i, y_i, z_i)\right]^2
    \label{eq:mse}
\end{equation}

\noindent where $N_P$ is the total number of interior points.

The CNN structure is composed of three dense layers with 128, 64, and 1 nodes, respectively. The Adaptive Moment Estimation (Adam) was used to optimize the weights of the dense layer during the network training, while the hyperbolic tangent was used as the activation function. We used a batch size equal to 32, and imposed a decaying learning rate, $\alpha$, given by Eq. \ref{eq:learning_rate}. A in depth description of the NLC-based CNN model is presented in \cite{aldeia2024temperature}.

\begin{equation}
    \alpha =
    \begin{cases} 
      10^{-3}, & \text{if } \text{epoch} < 300, \\
      10^{-4}, & \text{if } 300 \leq \text{epoch} < 600, \\
      10^{-5}, & \text{if } 600 \leq \text{epoch} < 900, \\
      10^{-6}, & \text{if } 900 \leq \text{epoch} < 1200 \\
    \end{cases}
    \label{eq:learning_rate}
\end{equation}

Through our computational model, we performed eleven simulations with varying peak linear heat generation rates ($q'_0$ in Eq. \ref{eq:linear_heat}) values to build the training, validating, and testing datasets. The case distribution between each of these sets is presented in Table \ref{tab:datasets}

\begin{table}[!h]
\centering
\caption{Distribution of cases between the training, validating and testing datasets (values in KW/m)
}
\label{tab:datasets}
\begin{tabular}{lc}
\hline
\textbf{Dataset}    & \textbf{Peak Linear Heat Generation Rate $\mathbf{q'_0\,\,\,\left[KW/m\right]}$} \\ \hline
\textbf{Training}   & 10, 12, 16, 18, 22, 24, 30, 36                                                   \\
\textbf{Validating} & 14, 16                                                                           \\
\textbf{Testing}    & 20                                                                               \\ \hline
\end{tabular}
\end{table}

\subsection{CNN Architecture Validation}

\begin{figure}[!ht]
    \centering \captionsetup{justification=centering}
    \includegraphics[width=\textwidth]{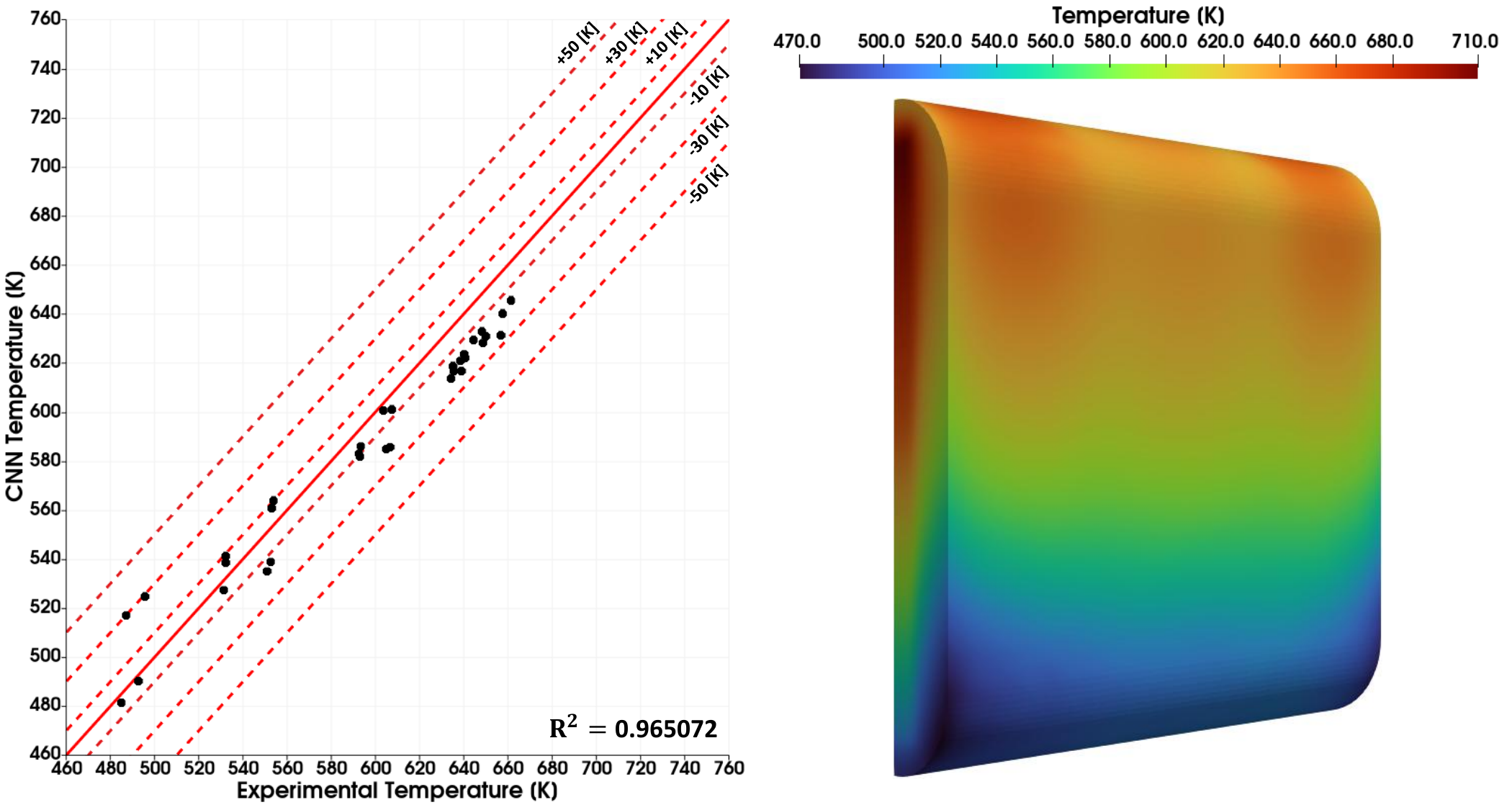}
    \caption{HTGR vessel wall temperature field reconstruction, adapted from \cite{aldeia2024temperature}.}
    \label{fig:cnn_validation}
\end{figure}

The validation of the CNN architecture was performed in \cite{aldeia2024temperature}, where they used the CNN model to perform the temperature field reconstruction over a metal plate representing a section of a high-temperature gas reactor vessel wall from a few thermocouple readings. Figure \ref{fig:cnn_validation} shows the HTGR vessel wall temperature distribution predicted by the NLC-based CNN architecture when experimental data was used as the input for the CNN. As we can see, the model achieved an $R^2$ above 0.96. 

\subsection{BISON/MOOSE-THM Coupled Model Verification and Validation}

\begin{figure}[!ht]
    \centering \captionsetup{justification=centering}
    \includegraphics[width=\textwidth]{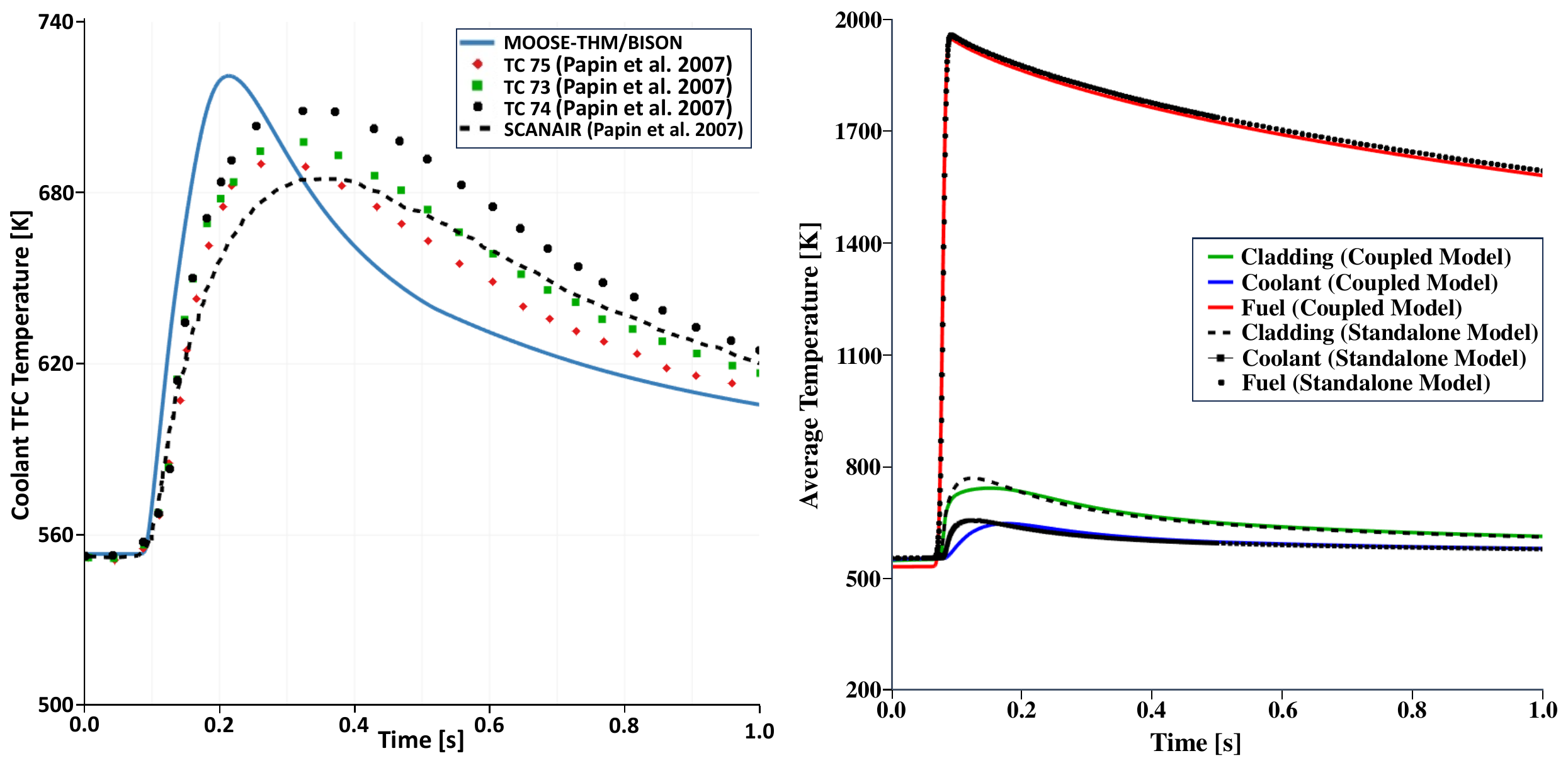}
    \caption{Coolant temperature validation (left) and average temperature code-to-code verification (right) of the CABRI Rep-Na3, adapated from \cite{machado2024toward}.}
    \label{fig:cabri_v_and_v}
\end{figure}

A fuel rodlet BISON/MOOSE-THM coupled model was validated in \cite{machado2024toward}. They compared the coolant temperature calculated by the coupled model against the experimental data presented in \cite{papin2007summary} for the CABRI Rep Na-3 test. They also performed a code-to-code verification between the coupled model and a BISON standalone model, comparing the average temperature in different regions of the fuel rodlet. Figure \ref{fig:cabri_v_and_v} present these two comparisons. 

\section{Results}
\label{sec:results}

This chapter presents the results of this work. We begin by introducing the predictions made by the NLC-based CNN model regarding the temperature distribution of the fuel rods. Next, we discuss the thermal-induced strain obtained from the predicted temperature distribution.

\subsection{CNN Temperature Distribution Prediction}

\begin{figure}[!ht]
    \centering \captionsetup{justification=centering}
    \includegraphics[width=\textwidth]{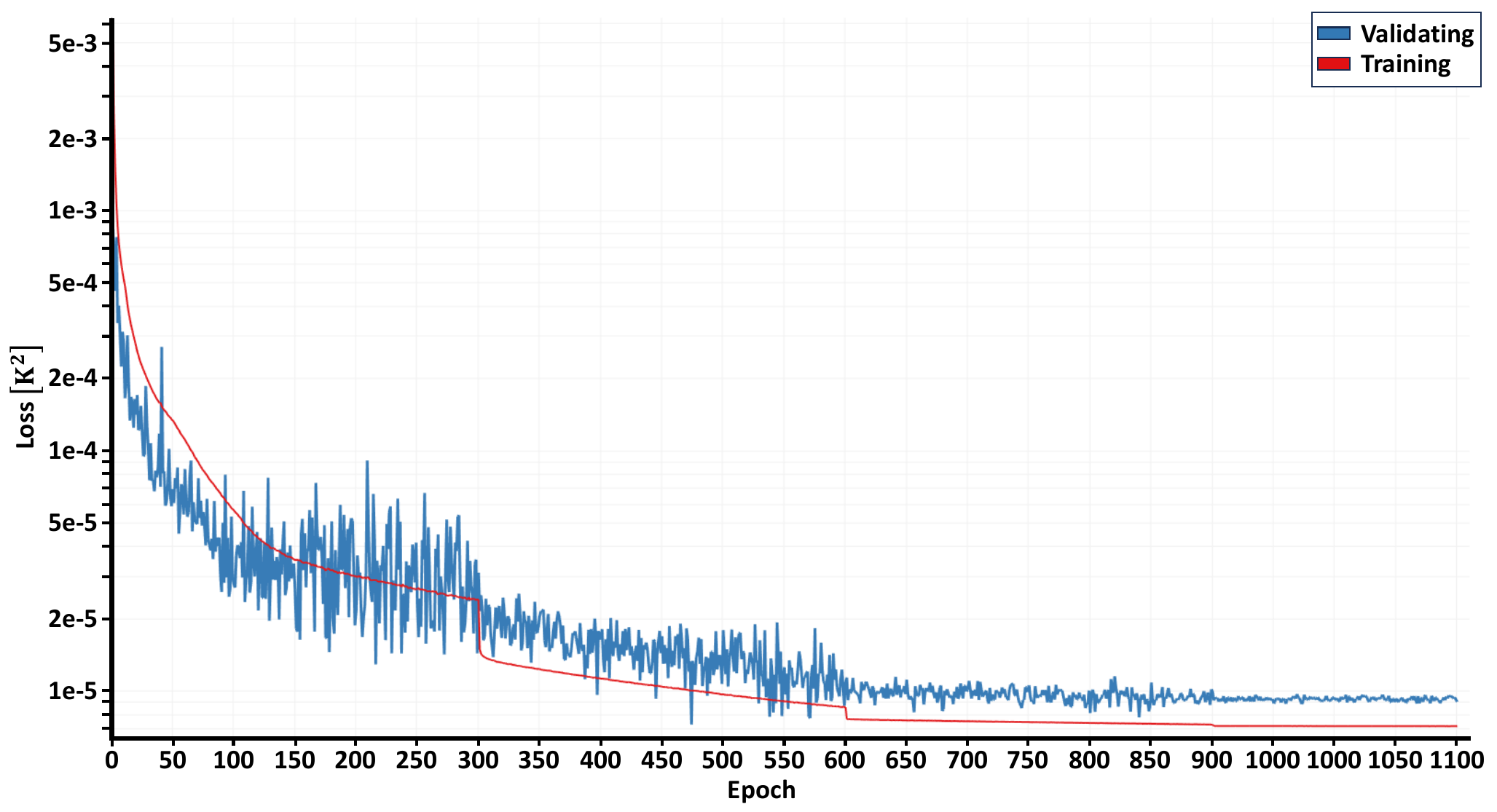}
    \caption{Loss vs Epoch curve.}
    \label{fig:loss_vs_epoch}
\end{figure}

The NLC-based CNN model was trained for 1,100 epochs without overfitting, as illustrated in Fig. \ref{fig:loss_vs_epoch}. After completing the training process, we employed the CNN to reconstruct the temperature field for a Pressurized Water Reactor (PWR) fuel rod with a peak linear heat generation rate of 20 $\text{[MW/m]}$. For this reconstruction, we assumed four known temperature measurements on the outer surface of the cladding, as depicted in Fig. \ref{fig:bison_geo}. Based on these measurements, we obtained the temperature distribution throughout the entire fuel rod domain.

\begin{figure}[!ht]
    \centering \captionsetup{justification=centering}
    \includegraphics[width=\textwidth]{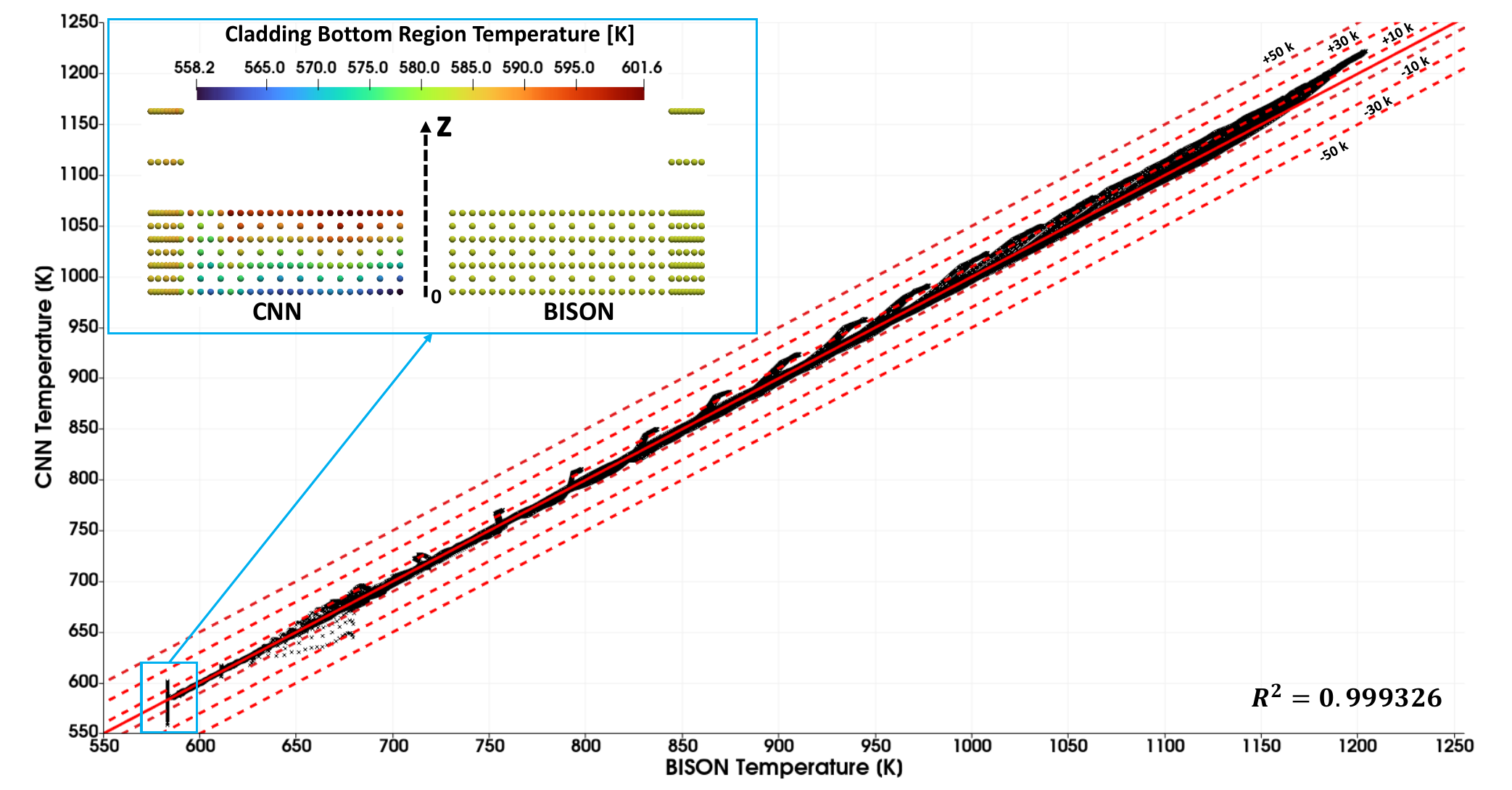}
    \caption{Temperature scatter plot for the test case.}
    \label{fig:scatter_plot}
\end{figure}

From Fig. \ref{fig:scatter_plot}, we can see that the CNN model successfully reproduced the temperature distribution for the fuel and cladding regions using only four temperature measurements collected from the cladding outer surface, achieving a coefficient of determination, $R^2$, above 0.99. 

This approach would enable estimating temperature across the entire fuel region using thermocouple readings from the cladding outer surface. This would, for example, allow the identification of undesired hot or cold regions within the fuel, which defective pellets may cause. In addition to the good agreement between the predicted and expected temperature values, we identified some interesting behaviors worth further investigation. 

The predicted temperature presented its higher deviation for the cooler regions of the fuel rod. In particular, our computational model calculated an almost constant temperature for the cladding bottom region, with values closer to the coolant inlet temperature, a behavior not reproduced by the CNN model, resulting in the near-vertical line that can be seen in the lower temperature regions of Fig. \ref{fig:scatter_plot}.

\begin{figure}[!ht]
    \centering \captionsetup{justification=centering}
    \includegraphics[width=\textwidth]{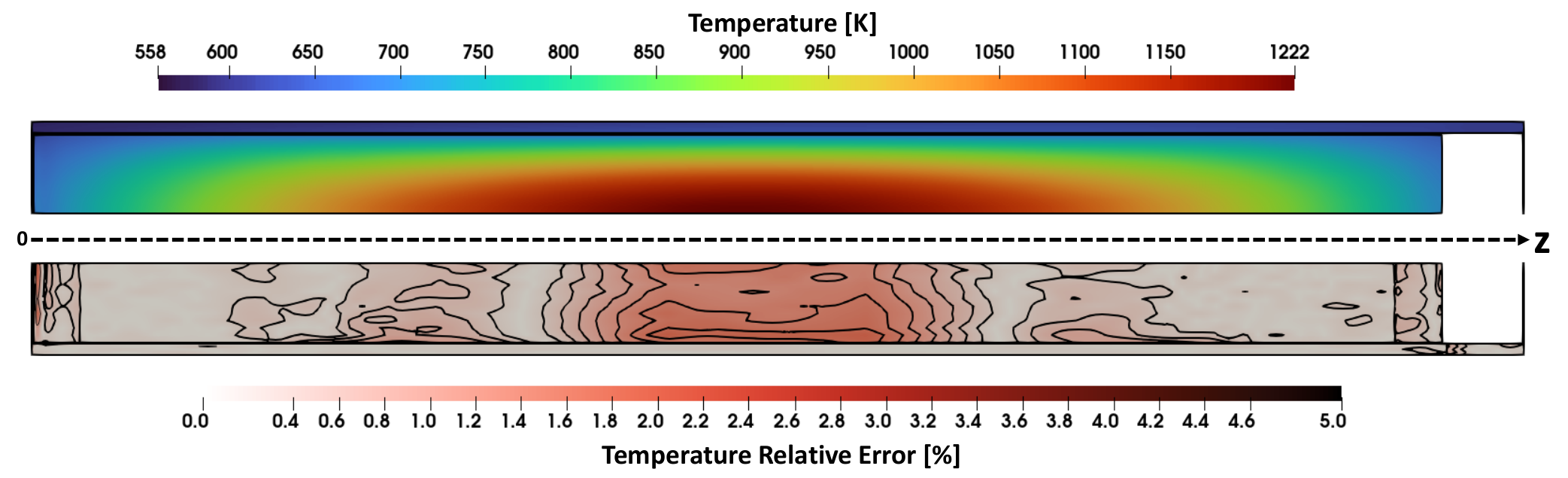}
    \caption{Predicted temperature and temperature relative error.}
    \label{fig:temp_contour}
\end{figure}

Figure \ref{fig:temp_contour} shows the reconstructed temperature distribution and the relative error between the computational model calculations and the CNN model prediction. As observed, the regions with the highest deviations are the fuel rod bottom, as mentioned earlier, and around the fuel rod axial center, where the highest temperatures occur. 

The reference model indicated that the gap between the fuel and cladding closed in this region thanks to the fuel expansion. This pellet-cladding mechanical interaction plays an important role in heat transfer between these two components and is believed to contribute to the higher deviation in the prediction for this region. However, further testing is necessary to confirm this hypothesis.

\subsection{Strain Estimation from Predicted Temperature Field}

As described in Sec. \ref{sec:thermomechanical}, we developed a simplified BISON model to analyze the thermomechanical response of the fuel rod. This model uses the predicted temperature distribution as input to calculate the total hoop strain and its components, including thermal-induced and creep hoop strains. Table \ref{tab:clad_strain} compares the results of this simplified model with those from the reference model presented in Sec. \ref{sec:thermomechanical}. It is worth mentioning that the simplified model considers only thermal creep in its creep hoop strain calculations, whereas the reference model also accounts for irradiation creep. 

\begin{table}[!h]
\centering
\caption{Cladding hoop strain comparison between the reference and simplified models}
\label{tab:clad_strain}
\begin{tabular}{lcc}
\hline
                                        & \textbf{Reference Model} & \textbf{Simplified Model} \\ \hline
\textbf{Thermal expansion hoop strain}  & 0.0021363                & 0.00214290                \\
\textbf{Creep hoop strain}              & 0.0001704                & 0.00005480                \\
\textbf{Elastic hoop strain}            & 0.0000465                & 0.00003700                \\
\textbf{Irradiation growth hoop strain} & -0.0003214               & 0.0000000                 \\
\textbf{Total hoop strain}              & 0.0020318                & 0.0022347                 \\
\textbf{run time [s]}               & 7005                     & 317                       \\ \hline
\end{tabular}
\end{table}

\begin{figure}[!ht]
    \centering \captionsetup{justification=centering}
    \includegraphics[width=\textwidth]{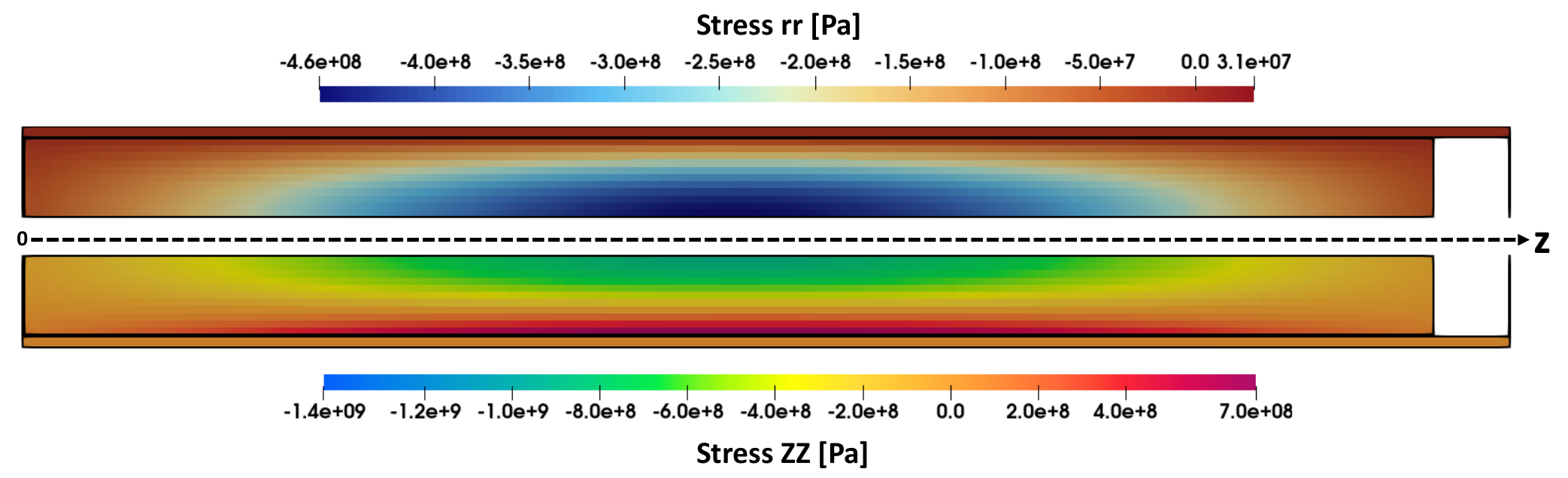}
    \caption{Estimated stresses in the z and r directions.}
    \label{fig:stress}
\end{figure}

The NLC-based CNN model combined with BISON allowed us to estimate important quantities for predicting possible structural failures, such as the total hoop strain. We could also estimate the stress distribution over the fuel rod in the axial and radial directions, as illustrated in Fig. \ref{fig:stress}. Thanks to the combined use of the CNN model and BISON, these estimations can be performed from thermocouple readings. If the relevant failure strains can be estimated using this method, it would support the predictive maintenance of nuclear reactors by allowing us to know the stress/strain of the fuel rod during operation. We are currently working on reducing the simplified thermomechanical model running time to provide a smaller window for stress/strain calculations, allowing real-time monitoring of such quantities.

\subsection{CNN Prediction Performance Across Fuel Burnup Levels}

To assess the performance of the CNN predictions regarding the fuel rod burnup level, we conducted a second set of tests where we simulated a total of 70 cases with burnups ranging from 2.4 to 59.7 $[MWd/kgU]$. From this set, 14 cases were randomly selected for validating, 14 for testing, and 42 for training the CNN. Due to the increased dataset size and associated computational cost, the CNN was trained for only 300 epochs using a fixed learning rate of (1e-3) in this test.

% \usepackage{graphicx}
\begin{table}[!h]
\centering
\caption{Coefficient of determination $\left(R^2\right)$ and normalized Euclidean norm $\left(NL^2\right)$ for temperature predictions performed at different burnups.}
\label{tab:burnup}
\resizebox{\textwidth}{!}{%
\begin{tabular}{lcccccccccccccc}
\hline
$\mathbf{Bu\,\,[MWd/kgU]}$ & \textbf{3.19} & \textbf{5.06} & \textbf{7.02} & \textbf{10.13} & \textbf{15.49} & \textbf{19.35} & \textbf{19.38} & \textbf{22.47} & \textbf{22.82} & \textbf{25.43} & \textbf{27.35} & \textbf{31.44} & \textbf{39.23} & \textbf{51.09} \\ \hline
$\mathbf{R^2}$             & 0.9829        & 0.9050        & 0.9598        & 0.9760         & 0.9151         & 0.9898         & 0.9611         & 0.9672         & 0.9761         & 0.9931         & 0.9865         & 0.9964         & 0.9890         & 0.9467         \\
$\mathbf{NL^2}$            & 0.0212        & 0.0421        & 0.0560        & 0.0358         & 0.0403         & 0.0167         & 0.0533         & 0.0466         & 0.0331         & 0.0274         & 0.0227         & 0.0194         & 0.0321         & 0.0868         \\ \hline
\end{tabular}%
}
\end{table}

Table \ref{tab:burnup} presents the coefficient of determination, $R^2$, and the normalized Euclidean norm for the 14 tests performed. As shown, even with the reduced number of training epochs, the predictions yielded an $R^2$ above 0.9 for all cases, with an average coefficient of determination across the test cases of approximately 0.96. The overall low normalized Euclidean norm values support the quality of the model predictions, though a few cases exhibit moderately higher errors. The sources of these discrepancies are still under investigation.

\section{conclusions}

In this work, we demonstrated a methodology that combines machine learning and computational models to estimate the temperature, stress, and strain states of a PWR fuel rod using a limited number of temperature measurements from the outer surface of the cladding.

We used the NLC-based CNN model to reconstruct the fuel rod temperature distribution from four temperature readings over the cladding outer surface. As illustrated in Fig. \ref{fig:scatter_plot}, the CNN model could perform the reconstruction with an $R^2$ above $0.99$. In a second test focused on determining the CNN performance for different burnup levels, the model achieved an $R^2$ above $0.9$ for all test cases, further demonstrating the model applicability for a broad range of operational conditions.

We also demonstrated that the NLC-based CNN architecture could perform good prediction even though, for the fuel rod, the cladding-coolant convective heat transfer coefficient varied as a function of the temperature. This indicates that such architecture could be used even when the HTC is not small and nearly constant, as stated by \cite{aldeia2024temperature}, extending its applicability.

By using the predicted temperature distribution as input for a thermomechanical model, we determined key parameters for assessing potential structural failures. Specifically, we presented the total hoop strain and the stress distribution along the axial and radial directions of the domain. This analysis highlighted the potential application of this methodology in the predictive maintenance of nuclear reactors, as thermocouple readings could be used to assess the fuel rod's state. This extends the findings of \cite{aldeia2024temperature}, where a similar study was conducted for the vessel wall of an HTGR.  

The simplified thermomechanical model calculated the maximum total hoop strain with a 10\% relative error compared to the reference model in less than 5\% of the time used by the reference model. However, further optimization is required to reduce its running time, allowing the simultaneous real-time estimation of the temperature field distribution and maximum hoop strain. 

Although the methodology produced accurate temperature predictions, it relied on temperature readings from the cladding surface, which are not available for commercial reactors. To extend the use of this methodology to such reactors we need to incorporate an additional step to calculate the fuel rod surface temperature using the available primary coolant temperature measurements. This calculated temperature can then be used to reconstruct the temperature across the fuel rod.

In cases where cladding temperature readings can be obtained, such as in experiments like the High-Burnup Experiments in Reactivity Initiated Accidents (HERA) \cite{kamerman2023high}, the methodology presented in this study could be directly applied to predict the fuel rod temperature distribution without further modifications.

It is worth mentioning that while important to enhance the reliability of this methodology, performing the uncertainty propagation through the CNN architecture is not trivial. As a future work we will focus on the development of methods to estimate the prediction uncertainty from input uncertainties. 

%Bibliography

\end{document}